\documentclass{article}

\usepackage[table,dvipsnames]{xcolor}
\usepackage{microtype}
\usepackage{graphicx}
\usepackage{subfigure}
\usepackage[most]{tcolorbox}
\usepackage{pdflscape}
\usepackage{adjustbox}
\usepackage{booktabs}
\usepackage{enumitem}
\usepackage{multirow}
\usepackage{hyperref}

\usepackage[accepted]{icml2025}

\usepackage{amsmath}
\usepackage{amssymb}
\usepackage{mathtools}
\usepackage{amsthm}

\usepackage{adjustbox}
\usepackage{multirow}
\usepackage{array}
\usepackage{makecell}  
\usepackage[capitalize,noabbrev]{cleveref}

\theoremstyle{plain}

\theoremstyle{definition}

\theoremstyle{remark}

\usepackage[textsize=tiny]{todonotes}

\icmltitlerunning{A Design and Evaluation Taxonomy for Natural Language Explanations}

\begin{document}

\twocolumn[
\icmltitle{A Taxonomy for Design and Evaluation of Prompt-Based Natural Language Explanations}




\begin{icmlauthorlist}
\icmlauthor{Isar Nejadgholi}{yyy}
\icmlauthor{Mona Omidyeganeh}{yyy}
\icmlauthor{Marc-Antoine Drouin}{yyy}
\icmlauthor{Jonathan Boisvert}{yyy}

\end{icmlauthorlist}

\icmlaffiliation{yyy}{National Research Council Canada, Canada}

\icmlcorrespondingauthor{Isar Nejadgholi}{isar.nejadgholi@nrc-cnrc.gc.ca}

\icmlkeywords{Machine Learning, ICML}

\vskip 0.3in
]



\printAffiliationsAndNotice{} 



\begin{abstract}


Effective AI governance requires structured approaches for stakeholders to access and verify AI system behavior. With the rise of large language models, Natural Language Explanations (NLEs) are now key to articulating model behavior, which necessitates a focused examination of their characteristics and governance implications. We draw on Explainable AI (XAI) literature to create an updated XAI taxonomy, adapted to prompt-based NLEs, across three dimensions: (1) \textit{Context}, including task, data, audience, and goals; (2) \textit{Generation and Presentation}, covering generation methods, inputs, interactivity, outputs, and forms; and (3) \textit{Evaluation}, focusing on content, presentation, and user-centered properties, as well as the setting of the evaluation. This taxonomy provides a framework for researchers, auditors, and policymakers to characterize, design, and enhance NLEs for transparent AI systems.

 
\end{abstract}

\section{Introduction}
\label{intro}

As AI is increasingly integrated into critical domains such as healthcare, finance, law, and defense, developing frameworks that bridge the gap between technical capabilities and governance requirements becomes essential \cite{reuel2024open}. The black-box nature of many AI models raises significant concerns around fairness, trust, and accountability \cite{Goodman2017}, and poses a barrier to effective governance by limiting access, and verification mechanisms. Explainable AI (XAI) directly addresses this opacity by providing methods and strategies that enhance transparency and make the decision-making processes of AI systems more understandable to humans through the use of \textit{explanations} \cite{das2020opportunities}. In general, explanations offer insights into how the AI system arrives at a specific decision and aim to empower certain stakeholders to \textit{access} and \textit{verify} the behavior of these systems effectively.

Traditionally, XAI has been more focused on making
the inner workings of deep learning models understandable to AI developers, with the goal of debugging and improving such systems \cite{minh2022explainable}. However, most of these explanations were not accessible to  other 
users, auditors or policymakers. With the emergence of LLMs, a unique opportunity arose in the field of XAI. Now, LLMs can be directly prompted and integrated into AI systems to provide textual explanations
which are referred to as prompt-based Natural Language Explanations (NLEs) \cite{costa2018automatic}. Such explanations serve as crucial interfaces between AI systems and human stakeholders due to their unprecedented fluency \cite{lakkaraju2022rethinking}.

While NLEs offer unique opportunities for enhancing AI transparency, they unfortunately introduce new challenges for governance due to their potential to present plausible yet unverified or misleading assertions. Despite many studies demonstrating successful human-AI collaboration through incorporating NLEs \cite{wiegreffe-etal-2022-reframing}, 
a growing body of research 
has shown that, if not carefully designed, these explanations can reinforce poor decisions. The limitations of NLEs include inconsistent or nonfactual explanations \cite{ye2022unreliability}, poor reasoning abilities  \cite{NEURIPS2023_ed3fea90} increase of over-reliance on the AI system \cite{zhang2020effect, wang2021explanations,poursabzi2021manipulating,bansal2021does,chen2023understanding}, and misleading outputs \cite{kayser2024fool}. 
This evidence calls for standardizing such explanations and creating guidelines that different stakeholders can adhere to when designing and validating such explanations. 

Specifically, the growing use of prompt-based NLEs necessitates a dedicated taxonomy that captures their distinct characteristics and governance implications. While the field of XAI has produced numerous taxonomies \cite{arrieta2020explainable, speith2022review}, none, to the best of our knowledge, are specifically tailored to NLEs. This work addresses that gap by adapting existing XAI taxonomies to the context of prompt-based NLEs, with the goal of informing their design and evaluation. Such a taxonomy advances technical AI governance by enabling the verification of model claims, providing meaningful access to system behavior for diverse stakeholders, and supporting the practical implementation of core principles such as accountability and transparency.

\section{Methodology and Scope}
\label{sec:method}
Our work builds on the XAI concepts synthesized in a recent meta-analysis by \citet{schwalbe2024comprehensive}, which consolidates insights from over 70 surveys into a unified taxonomy of methods, traits, and evaluation metrics. We refer to this tool as the XAI meta-taxonomy and adapt it to the specific case of NLEs.  A visual representation of the XAI meta-taxonomy is shown in Figure \ref{fig:taxonomy}, Appendix \ref{sec:appendix_meta}. For adaptation, we incorporate NLE-specific considerations from relevant literature, such as \citet{cambria2023survey}, who proposed a roadmap to understanding NLEs. We present our arguments for this adaptation and the related literature in section \ref{sec:taxonomy}. Here, we clarify the definition of NLEs, the scope of our study and our adaptation approach. 

\noindent \textbf{Definition of NLEs}: Inspired by \citet{cambria2023survey} and \citet{ ribeiro2016should}, we define NLEs as textual artifacts that \textbf{convey} a qualitative understanding of how an \textbf{input’s components }
influence a model’s prediction, along with \textbf{contextual information} relevant to a specific user or the task in general. This definition distinguishes NLEs from raw model reasoning traces or other outputs that are not centred explicitly around communication with the user.

\noindent \textbf{NLEs within the XAI landscape:} \citet{schwalbe2024comprehensive} synthesize the
XAI literature and present relevant concepts under 
three high-level procedural components: \textit{Problem Definition}, \textit{Explanator}, and \textit{Metrics}, with multiple hierarchical sub categorizations. While these components remain highly relevant for NLEs in general, their detailed subcategories require adaptation for the case of NLEs. Some elements are less directly applicable (e.g., formal mathematical constraints). Other subcategories demand greater nuance (e.g., interactivity in dialogue-based systems) or require more specificity in definitions.

First, while in general, a generative model might be fine-tuned or adapted otherwise to generate specialized explanations \cite{sammani2022nlx}, we limit the scope of this study to \textit{prompt-based} NLEs. Next, we narrow the scope of our work to a common class of prompt-based NLEs that are: (1) \textit{post-hoc} -- generated after the model has produced a decision \cite{retzlaff2024post}; (2) \textit{model-agnostic} -- not reliant on the internal architecture or parameters of the model \cite{ribeiro2018anchors}; and (3) \textit{local} -- produced with respect to individual instances rather than the model as a whole \cite{lundberg2020local}. Within this scope, we exclude components of XAI that pertain to inherently interpretable, self-explaining models or model-specific explanation techniques \cite{rudin2019stop}, and \textit{global} explanations that describe a model’s overall behavior \cite{saleem2022explaining}.

\noindent \textbf{Methodological approach:} Our adaptation of the XAI meta-taxonomy was informed by an iterative, consensus-driven process involving experts from multiple domains, including the authors. Specifically, we engaged stakeholders, including a researcher in explainable AI, several model developers, a technical manager, and an organizational decision-maker. Through a series of collaborative reviews, we identified which components of the original taxonomy align with the properties of prompt-based NLEs and where extensions or refinements were necessary.

\section{Proposed Taxonomy of Prompt-Based NLE}
\label{sec:taxonomy}
Table \ref{tab:taxonomy} presents our proposed taxonomy for post-hoc, model-agnostic, and local prompt-based NLEs, adapted from the meta-taxonomy (Figure \ref{fig:taxonomy}). We elaborate on each component and its nested subcategories within a two-tier structure. 

\begin{table*}[ht]
\scriptsize
    \centering
    \caption{Taxonomy for post-hoc, model-agnostic, and local prompt-based NLEs, adapted from existing XAI literature.}
    \label{tab:taxonomy}
    \renewcommand{\arraystretch}{1.2}
    \rowcolors{2}{gray!10}{white}
   \begin{adjustbox}{max width=\linewidth}
    \begin{tabular}{p{2.8cm} p{3.2cm} p{8.8cm}}
        \toprule
        \textbf{Component} & \textbf{Subcategory} & \textbf{Description/Example} \\
        \midrule
        \textbf{Context Definition} 
            & Task Types & Classification, regression, clustering, generation, etc. \\
            & Data Types & Structured (e.g., tables), unstructured (e.g., text, images, graphs). \\
            & Audience & Creator, Operator, Executor, Decision Subject, Examiner  \\
            & Explanation Goal & Model reasoning, task-level justification, or error detection\\
        \textbf{Generation \& Presentation} 
            & Model Type & Generative LLMs, VLMs. \\
            & Input & Prompt and context artifacts provided by users. \\
            & Interactivity & Dialogue-based refinement of explanation scope and relevance. \\
            & Output Type & Textual formats: examples, counterfactuals, structured narratives. \\
            & Presentation Form & Plain text, visuals, interactive widgets; adapted to user expertise and needs. \\
        \textbf{Evaluation} 
            & Content Properties 
            & Correctness, Completeness, Consistency, Continuity, Contrastivity, Comprehensibility\\
            & Presentation Properties 
            & Compactness, Composition, Confidence, Translucence \\
            & User-Centered Properties 
            & Actionability, Personalization, Coherence, Controllability, Novelty \\
            & Evaluation setting 
            & Functionally Grounded (automated),  Human-Grounded (subjective ratings), and Application-Grounded (performance in real-world tasks). \\
        \bottomrule
    \end{tabular}
    \end{adjustbox}
\end{table*}

\subsection{Context Definition}

\textit{Context Definition} is the first component of designing NLEs. This element corresponds to the \textit{Problem Definition} phase of the XAI meat-taxonomy, which aims to formulate the requirements of the task and the use case aspect. The subcategories of \textit{ Task Type} and \textit{Data Type} remain critically relevant to the case of NLEs. The former recognizes that the nature of the explanation depends on the underlying AI task, such as classification (explaining category assignment), Regression (justifying numerical predictions), or other tasks such as  detection. The latter differentiates between explanations for models trained on structured data (e.g., tabular data with defined variables) versus unstructured data (e.g., text, images, graphs), acknowledging that the form and content of explanations may differ significantly. Explanations for graph neural networks, for instance, present unique challenges \cite{yuan2022explainability}. 

As explained in Section~\ref{sec:method}, we omit the interpretability subcategory that pertains to explaining the intertwined process of prediction and explanation when they are performed jointly \cite{du2019techniques}. This exclusion is justified by the fact that NLEs are typically employed as \textit{post-hoc} methods, i.e., they are generated \textit{after} the model has made a decision. 

Under the \textit{Context Definition}, we add a new subcategory, \textit{Audience}, which is crucial for NLE but not well-represented in the XAI meta-taxonomy. Due to the versatility of NLEs, they can be generated for different stakeholders, and therefore, considering the audience is a major aspect of the \textit{context definition} for NLE. Here, we consider the following five categories of agents, proposed by \cite{tomsett2018interpretable}, who might be the audience of NLE: \textbf{1) Creator}: Agents who create the system, including owners and developers, and might need NLE for debugging, error analysis, etc., \textbf{2) Operator}: Agents who directly interact with the AI by providing the system with inputs and directly receiving the system’s outputs and might be interested in simple and informative NLEs to interpret the output or detect the model's failures. \textbf{3) Executors}: Agents who make decisions based on the output of the system and might be interested in contextual information relevant to the decision they want to make. Domain-specific jargon might be acceptable or even preferred. \textbf{4) Decision Subject}: Agents whose data is used in the training of the model, 
commonly interested in knowing how the system is using their data. \textbf{5) Examiners}: Agents who audit or investigate the model, potentially interested in explanations that describe why some rules or regulations have been violated or complied with. 

Moreover, we added the subcategory of \textit{Explanation Goal} to the \textit{context definition} category as it can vary 
based on the use case and audience. 
For this, we adopt the golas of NLE generation proposed by \cite{chen2022machine}, including: \textbf{1) Human Understanding of the AI Model:} Explain the model’s decision boundary 
so that humans can understand the model's reasoning. \textbf{2) Informing Human About the Task:} Explain the decision boundary of the task itself 
to enable humans to make better decisions 
 and \textbf{3) Enabling Users to Detect Model Errors:} Provide contextual, counterfactual, or confidence-level information to help users identify potential model errors.
 

\subsection{Generation and Presentation}
We rename the \textit{Explanator} module of the XAI meta-taxonomy to \textit{Generation and Presentation} of explanations in Table \ref{tab:taxonomy}. This component concerns how the explanation is generated and presented to users.  
Within the subcategories of this component, mathematical constraints like linearity or monotonicity are less critical for NLEs, as explanations are delivered in natural language rather than structured mathematical formats (e.g., decision rules or equations). For example, while a decision tree might require sparsity constraints, an NLE is usually agnostic to such constraints.  We also omit the \textit{portability} and \textit{locality} properties as we are focused on specific cases of model-agnostic and local (individual instance level) explanations.

Determining the \textit{explanator model type}, i.e., the model that is prompted to generate the explanations, is the first step. This can be one of the large families of generative models, including large language models \cite{kumar2024large} or vision language models \cite{zhang2024vision}. 
\textit{Input} to the explanator is another subcategory where the designer of NLE identifies the user prompt and other artifacts that are fed to the generative model. Furthermore, \textit{interactivity} is a lot more feasible in the case of NLE compared to traditional XAI models, as users can engage in dynamic dialogues to refine explanations. 
However, the line between \textit{interactivity} and required \textit{input}, is blurred in the case of NLE since user feedback, data, and context can all be provided through user interactions in the form of prompts. Incorporating insights from the field of human-computer interaction is essential in optimizing the interplay between interactivity and input types in prompt-based NLE \cite{reddy2024human}. 


Turning to the \textit{output type} subcategory, NLE outputs can range from concise summaries to structured narratives, depending on context. 
This depends on the explanation goal and the desired NLE's characteristics (see section~\ref{subsec:metrics}).

For the \textit{presentation} subcategory, the presentation of NLEs is highly adaptable and can be tailored to the user’s expertise and goals. For example, technical users may receive explanations with statistical terms (e.g., the confidence score), and laypersons might prefer simplified analogies (e.g., flagged transaction to spot unusual activity). Also, while the text is the primary presentation format, NLEs can integrate visual aids such as highlighting parts of input images \cite{kayser2024fool}, or interactive widgets such as sliders to explore ``what-if'' scenarios \cite{meyer2024slide}.

\subsection{Evaluation}
\label{subsec:metrics}

This component evaluates the quality of explanations (regardless of the predictive performance). Evaluating explanations is inherently complex due to the subjectivity of what a ``good'' explanation is, its dependence on context, the absence of a universally accepted ground truth, and the many desirable characteristics of an ideal explanation. Also, in XAI, a central tension exists between faithfulness (accuracy in reflecting the reasoning) and plausibility (being perceived as reasonable or satisfying by humans) \cite{jacovi-goldberg-2020-towards}. Prompt-based NLEs tend to favor plausibility over strict faithfulness, which comes at the cost of potentially oversimplifying or misrepresenting model internals.

In order to identify the potential desired characteristics of NLEs, we start with a framework with twelve explanation properties, Co-12, originally proposed by \citet{nauta2023anecdotal}. These properties are organized under three categories \textit{Content}, \textit{Presentation}, and \textit{User-centered} properties. The \textit{Content} category comprises: 1) Correctness, 2) Completeness, 3) Consistency, 4) Continuity, 5) Contrastivity, and 6) Covariate Complexity, and mainly focuses on the faithfulness of the explanation with respect to the model or task. \textit{Presentation} properties include: 7) Compactness, 8) Composition, 9) Confidence concern how explanations are formatted and structured.  Finally, \textit{User-centered} properties, include: 10) Context, 11) Coherence, and 12) Controllability, and emphasize the relevance and utility of explanations for end users. \textit{Presentation} and \textit{User-centered} properties generally fall under the plausibility criteria. 

Table \ref{tab:explanation-properties} presents 15 potential characteristics of NLEs, along with their descriptions, adapted from the above-mentioned Co-12 properties. Leveraging the usage-context-driven evaluation framework of \citet{liao2022connecting}, we adapt  the Co-12 framework to capture the unique affordances of NLEs. 
 Specifically,  1) We rename \textit{Covariate Complexity} as \textit{Comprehensibility}, aligning with the terminology of \citet{liao2022connecting}, which is more intuitive for language-based explanations. 2) We decompose the original \textit{Context} property (under the \textit{User-centered dimension}) into two distinct criteria, \textit{Personalization} (tailoring to user needs) and \textit{Actionability} (supporting decision-making), given the increased contextual adaptability of NLEs. 3) We add \textit{Translucence} to the \textit{Presentation} category to reflect an NLE’s ability to communicate its own limitations or conditions of validity. 4) We include \textit{Novelty} under \textit{User-centered} properties, recognizing that NLEs can provide new or surprising information derived from pre-trained knowledge. 

These properties might be evaluated in three settings: \textit{Application Grounded}, \textit{Human-Grounded} and \textit{Functionally Grounded} evaluations \cite{doshi2017towards}. Application-grounded evaluation credits explanations that improve the task in the real-world application and tests the explanations in the application context. While reaching high application-grounded results is the ultimate goal in generating explanations, designing and conducting such evaluations is resource-intensive and requires highly skilled users. Human-grounded evaluations offer an intermediate way of anticipating how the explanations would work in the real world by running human studies with laypeople. While human-grounded evaluations offer valuable insights into human perception, they can still be time-consuming. Functionally grounded evaluations provide alternative approaches that can be automated and do not require human.

\begin{table}[ht]
\centering
\scriptsize
\renewcommand{\arraystretch}{1.2}
\caption{Potential Desirable Explanation Properties}
\begin{tabular}{p{0.03cm} p{1.5cm} p{5.3cm}}  
\toprule
\textbf{\rotatebox{90}{}} & \textbf{Property} & \textbf{Description} \\
\midrule
\multirow{6}{*}{\rotatebox{90}{Content}} 
& Correctness & Accurately reflects AI's reasoning. \\
& Completeness & Covers all relevant factors contributing to the output. \\
& Consistency & Produces identical explanations for identical inputs. \\
& Continuity & Small input change results in small explanation change. \\
& Contrastivity & Highlights differences from alternative outcomes. \\
& Comprehensibility & Uses human-understandable concepts and relations. \\
\midrule
\multirow{4}{*}{\rotatebox{90}{Presentation}} 
& Compactness & Provides succinct and non-redundant explanations. \\
& Composition & Presents information clearly and structurally. \\
& Confidence & Communicates the model’s certainty or uncertainty. \\
& Translucence & Acknowledges explanation limitations or alternatives. \\
\midrule
\multirow{5}{*}{\rotatebox{90}{User-centered}} 
& Actionability & Supports the user in deciding what to do next. \\
& Personalization & Tailored to the user’s context or needs. \\
& Coherence & Aligns with user expectations and prior knowledge. \\
& Controllability & Enables user interaction or refinement of explanations. \\
& Novelty & Provides new or non-obvious information to the user. \\
\bottomrule
\end{tabular}
\label{tab:explanation-properties}
\end{table}

\vspace{-15pt}

\section{Example Use Case}

In this section, we elaborate on the application of the proposed taxonomy in ``anomaly detection in remote sensing using airborne imagery''. This task is crucial for public safety, border security, intelligence, surveillance, and reconnaissance, as well as natural disaster response. Anomalies include unusual or unexpected events, which signal important safety risks or security threats, such as ``a bus travelling on a bike path'' or a ``person standing in the middle of a roadway''. An effective AI-based system is not only expected to accurately detect such abnormalities but also to explain their nuances, to enable better-informed decision-making. 

Figure \ref{fig:system-prompt-example} highlights an example where a vision-language-based system is deployed to detect anomalies from images taken by a drone and communicate the necessary information to an operator. 
 In this setting, the operator requires clear and reliable explanations that justify why the system identified a particular behavior as abnormal. The taxonomy supports this need by setting the \textit{Explanation Goal} to clarify ``\textit{why the behavior deviates from normal patterns}''. It also guides the system to present the explanation in a structured format including 1) a binary decision (anomaly or not), 2) contrast between expected and observed behavior, and 3) a confidence level. This systematized prompting makes the explanation more straightforward to interpret, use and evaluate. For evaluation, Correctness, Confidence and Actionability of NLEs can be assessed in an application-grounded setting. 

The taxonomy also recognizes that different users need different explanations:
For developers, it is crucial to identify errors, such as false alarms, and analyze the model’s decision-making process for debugging purposes. Therefore, for this audience the goal of explanations shifts from ``\textit{Why the behavior deviates from normal patterns?}'' (as it is for the operator) to ``\textit{Why/how does the model detect this behavior as an anomaly?''}. The explanation may be presented as highlighted regions in the image or as counterfactual features showing what would change the model’s decision.

On the other hand, executors, such as field agents or decision-makers, need contextual and actionable explanations to understand the anomaly. Therefore, for them, the goal of explanations shifts to \textit{``What is the situation, and how should I respond?''} The response may include concise summaries of the anomaly, its potential implications, and recommended next steps to enable rapid and effective action in the field. Finally, examiners, such as auditors or policy reviewers, need explanations that demonstrate the system’s compliance with regulations, ethical standards, and operational guidelines. For examiners, the explanation goal becomes \textit{``Does this decision align with policy and was it reached in a valid way?''}. The response may take the form of a justification that includes traceable reasoning steps, references to decision criteria, and documentation of relevant inputs to support regulatory review and accountability.

\section{Discussion}
While the presented taxonomy provides a general framework for designing prompt-based NLEs, the most relevant elements should be empirically identified for each task with the following critical considerations in mind.

\noindent \textbf{Faithfulness as a necessity}: Although we mostly focused on the communicative aspect of NLEs, we recognize that, regardless of audience or context, explanations that distort, oversimplify, or obscure the system’s actual workings risk undermining transparency, accountability, and trust. Effective explanation systems must therefore not only adapt to the user and context, but also remain internally consistent and grounded in the model's underlying principles. In some cases, selective disclosure may be appropriate to accommodate user goals or protect sensitive information, but this must not compromise factual accuracy. Guaranteeing faithfulness is a challenging technical task that is beyond the scope of this study and deserves special attention \cite{yeh2019fidelity, lyu2024towards}.

\noindent \textbf{Subjectivity of Criteria}: Identifying an optimum prompt requires a comparative analysis of several settings based on the taxonomy, with respect to the desired criteria. However, many of the desired criteria we mentioned are highly subjective, which makes their evaluation particularly challenging. One effective paradigm in early design stages is to use the LLM-as-a-judge testing \cite{gu2024survey}, for automatic and cost-effective preliminary assessment. This approach involves using more advanced LLMs to evaluate the characteristics of the explanations generated by a production-grade model, and is suitable for early refinements before investing in high-cost, multi-annotator evaluations. However, care should be taken in such evaluations due to the limitations of LLMs when deployed as judges \cite{szymanski2025limitations}. 

\noindent \textbf{Trade-offs and conflicts } Crucially, there is a fundamental tension between some of the desired characteristics of NLEs shown in table \ref{tab:explanation-properties}. 
For example, achieving perfect \textit{Correctness} might result in an explanation that is highly complex and
detailed. Such an explanation, while greatly faithful and informative, could be difficult for a non-expert user to understand, and scores
low on \textit{Comprehensibility}. The ideal balance between these
properties often depends on the specific application context
and the needs of the user, especially in high-stakes, time-sensitive settings.

Specifically, AI assistance in sensitive or time-pressured environments may pose conflicts and risks that require striking a critical balance. For example, \citet{swaroop2024accuracy} found trade-offs between decision accuracy and speed, where simpler explanations led to faster decisions but increased the risk of overreliance on the AI, while more detailed explanations improved understanding but slowed down the response. These effects are further influenced by individual user traits, such as their baseline trust in AI and need to be mitigated relative to context, user goals, and cognitive constraints.

\section{Conclusion}
\label{conclusion}

Our taxonomy brings together key concepts relevant to NLEs and offers a practical checklist for designers and evaluators while supporting a systematic approach to AI governance. It aligns with several capacities outlined by \citet{reuel2024open}, including: 
access, by enabling interaction through natural language without needing internal model access; and verification, by guiding the design of explanations that can substantiate claims about system behavior.
Moreover, this taxonomy could support the operationalization of transparency and accountability by aligning explanation strategies with specific audiences and objectives. Future work involves validating its use through experiments in collaboration with human-computer interaction experts.


\section*{Acknowledgment}
 This work was conducted by the National Research Council Canada on behalf of the Canadian AI Safety Institute. 

\bibliography{references}
\bibliographystyle{icml2025}


\appendix
\onecolumn

\section{Meta-Taxonomy of XAI}
\label{sec:appendix_meta}
Figure \ref{fig:taxonomy} presents the meta-taxonomy, whose components we systematically analyzed to develop our proposed taxonomy.

\setcounter{figure}{0}
\renewcommand\thefigure{A.\arabic{figure}} 

\begin{figure}[H]
    \centering
    \includegraphics[width=0.9\linewidth,clip,trim=0cm 0cm 3cm 0cm]{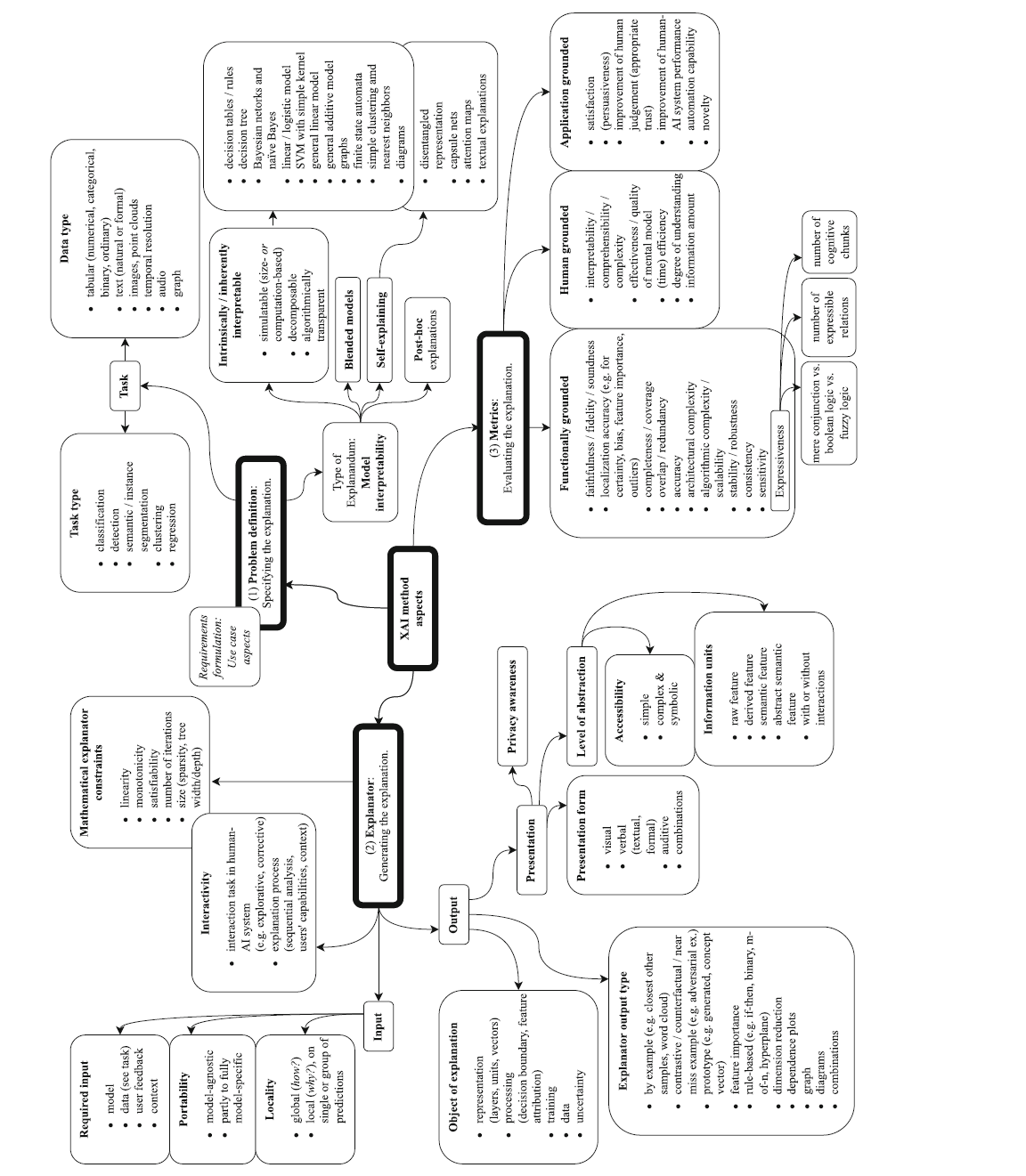}
    \caption{The XAI meta-taxonomy proposed by \citet{schwalbe2024comprehensive} (Page 50, Figure 7).}
    \label{fig:taxonomy}
\end{figure}

\section{Example Prompt Guided by Taxonomy}
\label{sec:example}
\setcounter{figure}{0}
\renewcommand\thefigure{B.\arabic{figure}} 
Here, we show an example explanation for the task of traffic anomaly detection guided by the taxonomy. 
\begin{figure}[H]
\centering
\small
\begin{minipage}{\textwidth}
\centering
\small
\includegraphics[width=0.15\textwidth]{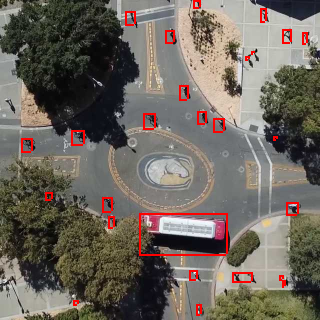}

\begin{tcolorbox}[title=SYSTEM PROMPT, coltitle=white, colbacktitle=blue!60!black]
\textbf{TASK TYPE:} \\
Your task is to detect and explain anomalies in the provided input. You carefully analyze the input image to detect irregularities or deviations from expected patterns or behaviors within the given image.

\vspace{0.6em}

\textbf{DATA TYPE:} \\
You work with images captured by a camera mounted on an aerial vehicle. Red boxes in the image represent moving objects. 

\vspace{0.6em}

\textbf{AUDIENCE:} \\
Your explanations will be used by a technical operator who manages the aerial vehicle.

\vspace{0.6em}

\textbf{EXPLANATION GOAL:} \\
In your explanation, you provide evidence for \textbf{WHY} the detected behaviour is an anomaly. Human operator uses your explanation to understand the nature of the anomaly.

\vspace{0.6em}

\textbf{OUTPUT TYPE:} \\
You generate clear, precise textual explanations that include:
   \vspace{-0.5em}
\begin{itemize}
    \item Binary anomaly detection decision
    \vspace{-0.5em}
    \item Description of both expected behaviors and observed deviations with contrastive argument.
    \vspace{-0.5em}
    \item A binary confidence level (high/low) for the identified anomaly.
\end{itemize}

\vspace{0.4em}

\textbf{PRESENTATION FORMAT:}
\vspace{-0.5em}
\begin{itemize}
    \item \textbf{Anomaly Detected:} This only includes YES or NO. YES refers to the presence of anomalies, and NO refers to normal cases. Do not include anything else.
    \vspace{-0.5em}
    \item \textbf{Expected Normal:} This includes a detailed description of typical behavior or pattern expected in the region.
     \vspace{-0.5em}
    \item \textbf{Observed Anomaly:} This includes a clear contrast between the observed deviation and the expected behavior.
     \vspace{-0.5em}
    \item \textbf{Confidence Level:} This states the confidence level regarding the presence of the anomaly in the image and includes "High" or "LOW, Manual Check Recommended". Do not include anything else.
\end{itemize}
\end{tcolorbox}


\begin{tcolorbox}[title=RESPONSE, coltitle=white, colbacktitle=blue!60!black]
\textbf{Anomaly Detected:} Yes

\vspace{0.6em}

\textbf{Expected Normal:} In a typical roundabout, vehicles are expected to navigate around the central island in a counter-clockwise direction, following the designated lanes and road markings. The roundabout is designed to facilitate smooth traffic flow, with vehicles entering and exiting at designated points without obstruction.

\vspace{0.6em}

\textbf{Observed Anomaly:} The image shows a bus positioned across the roundabout, blocking the flow of traffic. This is a deviation from the expected behavior as the bus is not following the roundabout's circular path and is instead obstructing the lanes, potentially causing a traffic jam and disrupting the normal flow of vehicles.

\vspace{0.6em}

\textbf{Confidence Level:} High
\end{tcolorbox}

\end{minipage}
\caption{A system prompt for the task of traffic anomaly detection guided by the XAI taxonomy proposed in Table \ref{tab:taxonomy} and the generated response by gpt-4o-2024-08-06 for the image shown above (without user prompt). Red boxes within the image represent moving objects. The image is sourced from \url{https://gitlab.lrz.de/ai4eo/reasoning/drone-anomaly} and red boxes are added through implementation of computer vision techniques.  
This dataset is originally used to evaluate existing anomaly detection methods and provides a benchmark for the task of training and testing video processing models. 
The authors would like to acknowledge Joshua Power and Delasie Torkornoo, Software Developer Specialists at NRC, for their contribution to producing the results presented in this appendix.
}
\label{fig:system-prompt-example}
\end{figure}

\end{document}